\documentclass{bmvc2k}

\usepackage{ctable}
\usepackage{multirow}
\usepackage{xcolor, colortbl}
\usepackage{comment}

\newcommand{\methodAcronym}{ASGRA}
\newcommand{\methodName}{\textbf{A}ttention over \textbf{S}cene \textbf{Gr}aphs for Sensitive Content \textbf{A}nalysis}

\title{Attention over Scene Graphs: Indoor Scene Representations Toward CSAI Classification}

\addauthor{Artur Barros}{artur.barros@students.ic.unicamp.br}{1}
\addauthor{Carlos Caetano}{caetanoc@unicamp.br}{1}
\addauthor{João Macedo}{joaomacedo@gmail.com}{2, 3}
\addauthor{Jefersson A. dos Santos}{j.santos@sheffield.ac.uk}{4}
\addauthor{Sandra Avila}{sandra@ic.unicamp.br}{1}

\addinstitution{
 Instituto de Computação (IC)\\
 Universidade Estadual de Campinas (UNICAMP)\\
 Campinas, São Paulo, Brazil
}
\addinstitution{
 Departamento de Ciência da Computação (DCC)\\
 Universidade Federal de Minas Gerais (UFMG)\\
 Belo Horizonte, Minas Gerais, Brazil
}
\addinstitution{
 Polícia Federal (PF)\\
 Belo Horizonte, Minas Gerais, Brazil
 }
\addinstitution{
 School of Computer Science\\
 University of Sheffield\\
 Sheffield, England, United Kingdom
}

\runninghead{Barros et al.}{Attention over Scene Graphs}

\begin{document}

\maketitle

\begin{abstract}
Indoor scene classification is a critical task in computer vision, with wide-ranging applications that go from robotics to sensitive content analysis, such as child sexual abuse imagery (CSAI) classification. The problem is particularly challenging due to the intricate relationships between objects and complex spatial layouts. In this work, we propose the \methodName~(\methodAcronym), a novel framework that operates on structured graph representations instead of raw pixels. By first converting images into Scene Graphs and then employing a Graph Attention Network for inference, \methodAcronym~directly models the interactions between a scene's components. 
This approach offers two key benefits: (i) inherent explainability via object and relationship identification, and (ii) privacy preservation, enabling model training without direct access to sensitive images. On Places8, we achieve 81.27\% balanced accuracy, surpassing image-based methods. Real-world CSAI evaluation with law enforcement yields 74.27\% balanced accuracy. Our results establish structured scene representations as a robust paradigm for indoor scene classification and CSAI classification. Code is publicly available at \url{https://github.com/tutuzeraa/ASGRA}.

\end{abstract}
\section{Introduction}

Scene classification is a fundamental problem in computer vision, especially for indoor environments~\cite{patel2020survey}. The goal is to categorize an image according to predefined scene types (e.g., bedroom, living room), with indoor scene classification focusing on interior spaces. Beyond general-purpose applications, indoor scene classification has also proven useful in highly sensitive domains, such as Child Sexual Abuse Imagery (CSAI) classification \cite{Coelho:2025,valois2025leveraging}. Prior studies involving interviews with law enforcement agents have shown that environmental and object-based contextual cues within a scene can serve as critical indicators of inappropriate content~\cite{Kloess:2021}. 

Despite the availability of powerful approaches for indoor scene classification~\cite{wang2023internimage}, significant limitations persist \cite{patel2020survey}. The task is particularly demanding due to the high complexity and variability of indoor scenes, characterized by a sheer diversity of objects, textures, and colors. Additionally, inherent ambiguity and inter-class similarity, such as the subtle distinction between a bedroom and a child's room, make this task particularly challenging. 

Scene Graphs (SGs)~\cite{Jung:CVPR:2023} offer a promising approach, representing scenes as structured graphs where objects are nodes and their relationships are edges. Each scene is modeled as a set of triplets in the form (\texttt{subject}, \texttt{predicate}, \texttt{object}), such as (\texttt{bed}, \texttt{next}~\texttt{to}, \texttt{window}), which explicitly encode semantic and spatial interactions between entities. Unlike traditional image representations, SGs explicitly encode semantic and spatial interactions between objects, offering a compact and interpretable abstraction of the scene. This structured format can be particularly valuable in indoor environments where object relationships distinguish visually similar scenes.

We introduce the \methodName~(\methodAcronym), a novel framework that leverages the inherent structure of the scene, modeled via SGs, to improve indoor scene classification. We use the extracted SGs as input to a Graph Attention Network~\cite{Velickovic2018GGraphAttentionNetworks} (GAT), which effectively weighs the importance of each triplet within the scene graph, thereby enhancing the model’s robustness in discerning similar classes. 

One of the primary motivations for adopting this pipeline is its suitability for handling highly sensitive content, such as CSAI. Since direct training on CSAI datasets is ethically prohibitive and legally constrained, our approach offers a practical solution: using only scene graph representations and corresponding high-level labels (e.g., ``CSAI'', ``Not CSAI'') from law enforcement agents. This enables effective model training while maintaining strict adherence to ethical and legal standards, without exposing researchers to harmful content.

\section{Related Work}

Pioneering work by \citet{Quattoni2009MITIndoorSR} showed that indoor scenes possess unique characteristics that make their classification inherently more difficult than outdoor environments. Consequently, specialized datasets such as MIT Indoor Scenes~\cite{Quattoni2009MITIndoorSR} and 
Places8~\cite{valois2025leveraging} emerged, promoting focused research on indoor contexts.

More recent approaches leverage deep convolutional neural networks to capture local and global features \cite{Sajjadi:2023:CVPR, Song:2023:IJCNN, Zhou2017PlacesSR, zhou2014}, while methods employing Graph Neural Networks (GNNs) explicitly model object relationships and spatial layout within scenes~\cite{chen2020scene, Qiu2021SceneEssence, Fan:2022:SRRN, Belmecheri:2025:Explainable}. In particular, these graph-based methods have demonstrated improved performance by encoding relational semantics and structural information explicitly, thereby effectively addressing ambiguities inherent to indoor scenes.

Most recently, \citet{valois2025leveraging} and \citet{coelho2024transformers} have developed complementary methods for scene classification, evaluated on Places8. \citet{valois2025leveraging} present a comprehensive study of self-supervised approaches, showing that a ResNet-50 fine-tuned on the Barlow Twins protocol can leverage unlabeled data to achieve strong performance. In contrast, \text{\citet{coelho2024transformers}} propose a few-shot learning framework based on Vision Transformers (ViTs), demonstrating competitive results with only five annotated examples per class. However, both approaches rely on image-based representations rather than structured graph representations, limiting their capacity to model complex spatial and semantic object relationships critical for disambiguating similar indoor scene categories.

\section{Our Approach}

Our problem is formulated as follows: given an arbitrary indoor image $I$, our objective is to classify it into one of the predefined indoor scene categories $y \in C$. Instead of operating directly on pixel data from the image $I$, we extract an SG representation, $G = (V, E)$, where $V$ is a set of nodes representing the objects and $E$ is a set of directed edges representing the relationships between them. After, we treat it as a graph classification problem: given the scene graph~$G$, the goal is to predict its scene label $y \in C$.

\begin{figure*}[t!]
   \centering
   \includegraphics[width=0.95\linewidth]
   {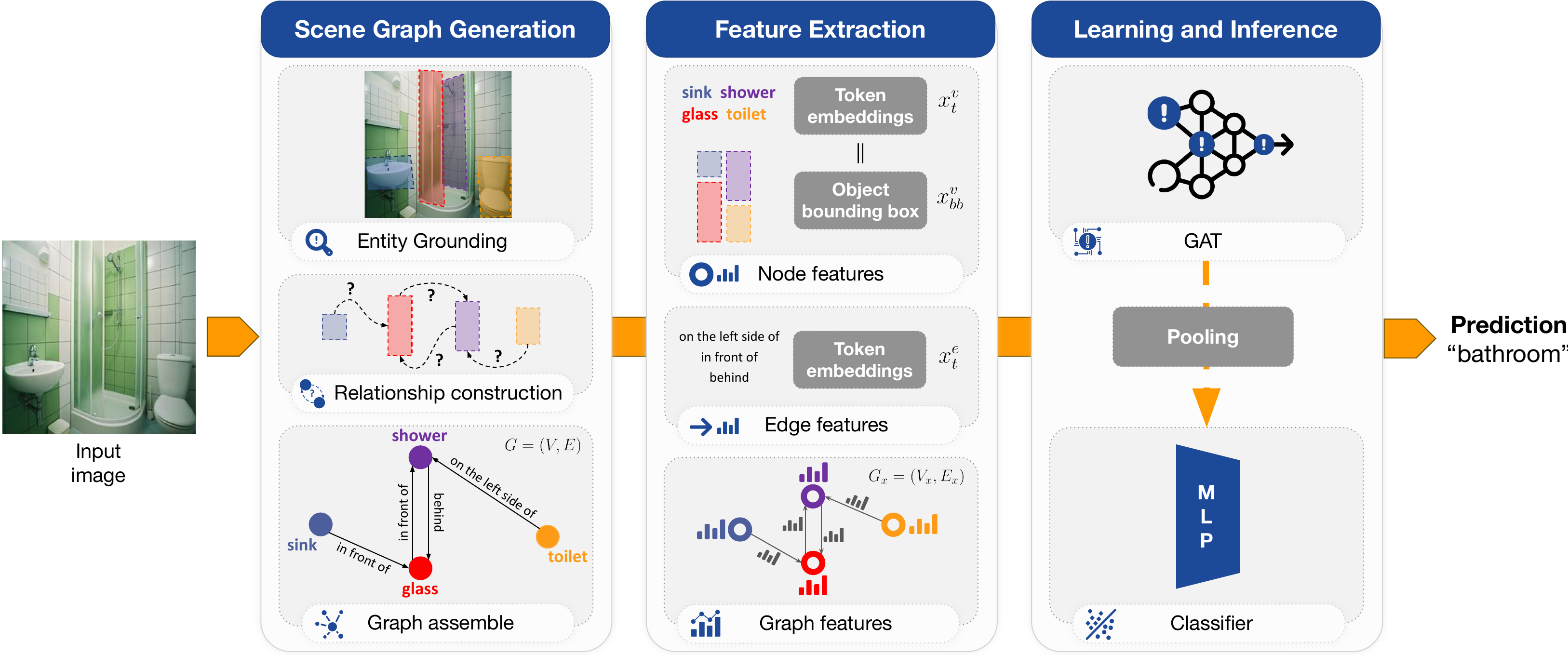}
   \caption{The \methodAcronym~framework processes input images through a pre-trained SGG model to generate structured graph representations. Detected objects and bounding boxes become node features while relations form edge features. A GAT performs learning and inference, with attention pooling, and a multilayer perceptron (MLP) predicts the indoor scene category.
   }
   \label{fig:sgsc_pipeline}
\end{figure*}

Our proposed \methodAcronym~framework consists of three main steps: (i)~scene graph generation, (ii)~feature extraction, and (iii)~learning and inference. Fig.~\ref{fig:sgsc_pipeline} illustrates our pipeline. 

For the first step, we leverage Pix2Grp~\cite{li2024pixels}, an off-the-shelf SGG model based on a Vision-Language Model architecture. Pix2Grp has an entity grounding module with the aim of predicting the bounding boxes and labels of the objects in the scene. With the objects in hand, a relationship construction module generates spatial and category labels to create the relation triplets. Finally, we can perform a graph assembly that creates the scene graph $G$.

In the feature extraction step, node features are constructed by concatenating two types of information for each node \text{$v \in V$: $x_t^v$}, derived from the token embeddings of the object labels; and $x_{bb}^v$, the normalized bounding box coordinates of the detected object. The final node representation is a concatenation of the aforementioned features: $x_{t \Vert bb}^v =$~$x_t^v$~$\Vert$~$x_{bb}^v$. For each edge $e \in E$, we extract features $x_t^e$, derived from the token embeddings of the predicate (relation). Finally, our graph features can be represented as $G_x = (V_x, E_x)$, where $V_x = \{x_{t \Vert bb}^0,$ $x_{t \Vert bb}^1, \dots , x_{t \Vert bb}^v\}$ is the set of node features and $E_x = \{x_{t}^0,$ $x_{t}^1, \dots , x_{t}^e\}$ is the set of edge features.

In the learning and inference step, we employ the GATv2~\cite{brody2021attentive}. In this step, the graph is processed by computing attention coefficients for each edge, dynamically weighting the influence of neighboring nodes during message passing. This attention mechanism allows the model to focus on the most relevant triplets that drive the prediction. Finally, a graph pooling layer aggregates the node representations into a single graph-level vector, which is passed to an MLP for the final scene classification.

\section{Experimental Results}

\subsection{Experimental Setup}

\noindent \textbf{Datasets.} We evaluate our benchmark on the Places8~dataset~\cite{valois2025leveraging}. Places8 is a curated subset of the Places365 dataset~\cite{Zhou2017PlacesSR}, consisting of 407,640 images (256$\times$256 pixels) selected from 23 of the original 365 scene classes. The authors remapped these classes into 8 indoor scene categories, chosen for their relevance to frequently encountered environments in CSAI. We follow the train/val/test experimental protocol proposed by the authors\footnote{Dataset splits are available at \url{https://doi.org/10.5281/zenodo.13910525}.}. 

For sensitive media evaluation, we collaborate with law enforcement to use the Region-based Annotated Child Pornography Dataset (RCPD)~\cite{Macedo2018BenchmarkMethodologyChild}, a private dataset maintained by the Brazilian Federal Police. Originally designed for forensic analysis, the dataset lacks~standard train/test splits and prohibits direct training access by non-law-enforcement personnel.

Through formal police collaboration, authorized agents perform SGG internally on RCPD images, providing only resulting scene graphs to our research team. This ensures no sensitive image access during the study. We train models using 5-fold cross-validation on these graph representations, enabling effective evaluation without direct CSAI data access.\vspace{0.25cm}

\noindent \textbf{Vision-Language Model Baseline.} To establish a solid point of comparison, we adopted a Vision-Language Model (VLM) configured for a Visual Question Answering (VQA) task, tailored to scene classification. VQA is a task in which a model receives an image alongside a textual question and must generate an appropriate answer, combining visual perception with natural language understanding. Our baseline uses the Large Language and Vision Assistant~(LLaVA)~\cite{Liu:2023:LLaVA}, which combines a vision encoder (e.g., CLIP~\cite{Radford:2021:CLIP}) with a language decoder (e.g.,~Vicuna~\cite{Chiang:2023:Vicuna}). LLaVA's multimodal instruction tuning enables highly accurate, context-aware responses, making it an effective choice for our scene recognition task.

One of the primary reasons for adopting a VLM in this role is its versatility. Trained on large-scale multimodal datasets, these models can adapt to a wide variety of visual contexts and maintain performance even in challenging conditions~\cite{Yue:2023:CVPR, Yue:2025:ACL}. Additionally, using language in the form of VQA provides a flexible and robust evaluation protocol \cite{Tong:2024:NeurIPS}. 

The overall pipeline operates in a simple yet effective manner. Images are processed individually, each paired with the corresponding textual prompt. The model's outputs are parsed to extract the predicted category, which is then compared with the ground truth labels for evaluation. This approach offers a robust and easily reproducible benchmark, enabling a fair assessment of more specialized methods introduced in later sections.\vspace{0.25cm}

\noindent \textbf{Scene Graph Generation and Feature Extraction.} We employ Pix2Grp~\cite{li2024pixels} to generate scene graphs due to its robust performance. The model outputs triplets comprising predicted subjects, objects, and their relationships. We used the model weights pre-trained on VG150~\cite{Krishna:IJCV:2017}, one of the most widely adopted datasets for evaluating SGG~\cite{Xu:CVPR:2017,Tang_2019_CVPR,li2024pixels}. Consequently, the predicted triplet labels are constrained to the 150 object classes and 50 relationship classes defined in VG150, which do not contain object classes (e.g.,  intimate body parts) or relationship classes (e.g., hugging, kissing, touching) directly related to CSAI. 

For feature representation, each node is encoded by concatenating its label index with the detected bounding box coordinates, while edge features correspond to the relation token ID. Importantly, we avoid explicit image features to ensure privacy preservation, as incorporating such features could enable reconstruction of sensitive images from trained models~\cite{Carlini2021ExtractingTrainingData}, which is undesirable for sensitive media applications.
\vspace{0.25cm}

\noindent \textbf{Implementation Details.} 
All experiments are conducted on 5 NVIDIA RTX5000 GPUs. Hyperparameter optimization is performed using the Optuna framework to efficiently explore optimal configurations.

For the baseline, we use LLaVA version 1.6 with the Vicuna decoder and provided hyperparameters. For our experiments, the model receives along the input image the following prompt: \textit{``Classify the received image into one of the following 8 categories: (0) bathroom; (1) bedroom; (2) child's room; (3) classroom; (4)~dressing room; (5) living room; (6) studio; or (7) swimming pool. Answer with only the number of the corresponding category provided.''} This prompt format ensures that predictions are both interpretable, constrained to the desired label set and easy to parse.

The GAT model is trained using early stopping with a patience of 10 epochs based on validation loss, with a maximum of 120 epochs. Cross-entropy loss is optimized via the Adam optimizer~\cite{Kingma2015ADAM}, with an initial learning rate of $1\times10^{-4}$ and weight decay of $3\times10^{-5}$. A batch size of 8 is used. To mitigate overfitting, a dropout rate of 0.2 is applied across all GATv2 layers. The final architecture comprises two GATv2 layers, 364 hidden dimensions, and 4 attention heads.

For the CSAI classification task, each fold of the 5-fold cross-validation protocol is trained for 20 epochs, with a learning rate of $3.8\times10^{-4}$, batch size of 8, and weight decay of $1.4\times10^{-5}$. The final CSAI model consists of two GATv2 layers, 128 hidden dimensions, and 8 attention heads. 

\subsection{Results and Analysis}

\noindent \textbf{Quantitative Analysis.} On Places8, the VQA-baseline achieved 77.69\% balanced accuracy on the test split, surpassing previous approaches: self-supervised learning (71.60\%~\cite{valois2025leveraging}) and few-shot learning (73.50\%~\cite{coelho2024transformers}). Despite being computationally heavy (7B parameters), this baseline establishes a strong comparison point. Our proposed \methodAcronym~achieved superior performance at 81.27\% balanced accuracy using only 242 million parameters, demonstrating efficiency and scalability. Table~\ref{tab:validation_results} presents detailed comparisons. 

\begin{figure}[t]
   \centering
    \begin{tabular}{cc}
    \includegraphics[width=0.48\linewidth]{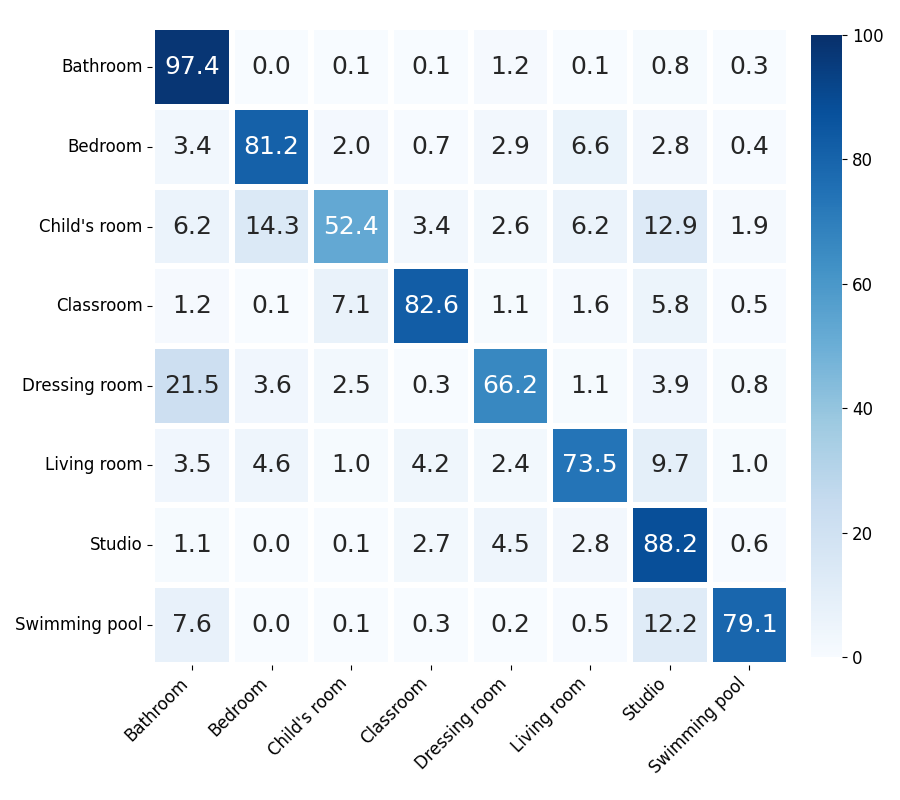} &
    \includegraphics[width=0.48\linewidth]{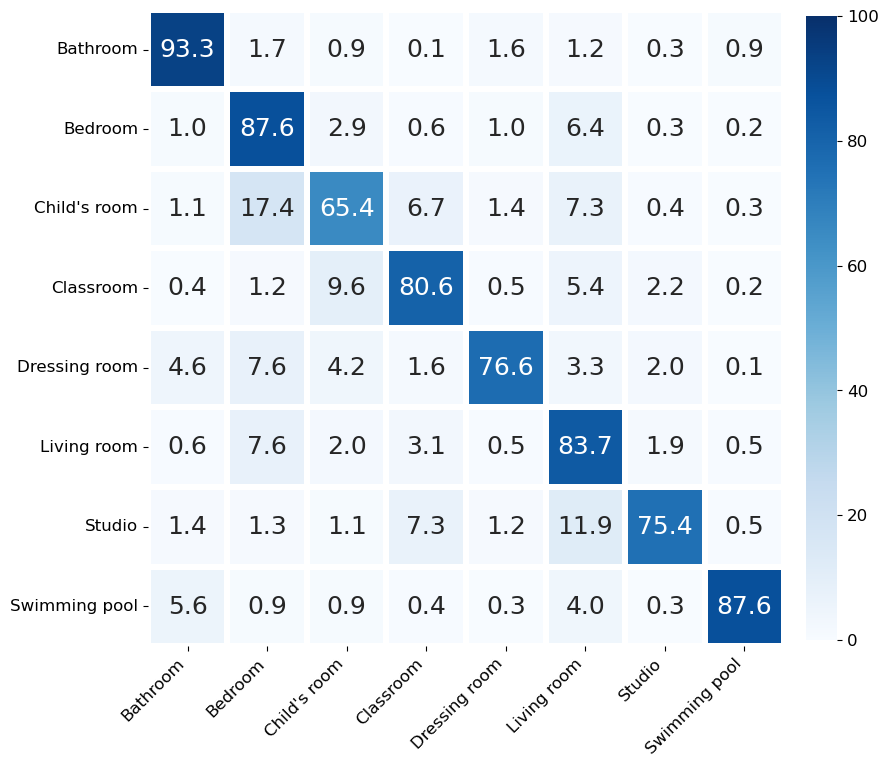} \\
    (a) VQA-baseline & (b) \methodAcronym \\\\
    \end{tabular}
   \caption{Confusion matrices on the Places8 test split.}
   \label{fig:confusion_matrices}
\end{figure}

\begin{table}[!ht]
\centering 
\footnotesize
\begin{tabular}{lcccc}
\midrule
\textbf{Method} & \textbf{Core Component(s)}  & \textbf{Input Modality} & \textbf{\#Params}  & \textbf{Acc. (\%)} \\
\midrule
VQA-baseline~\cite{Liu:2024:LLaVA1.6-NeXT} & LLaVA~v1.6 + Vicuna & Image \& Text & 7B & 77.69\\
Few-shot \cite{coelho2024transformers}      & ViT-Small & Image & 21.7M & 73.50 \\
Self-supervised \cite{valois2025leveraging} & ResNet-50 & Image & 23.5M & 71.60 \\
\methodAcronym~(ours) & Pix2Grp + GATv2 & Scene Graph & 242M & \textbf{81.27} \\
\midrule
\end{tabular}
\caption{Comparative results on the Places8 test set. Our \methodAcronym~framework is benchmarked against state-of-the-art methods. 
\methodAcronym~operates on scene graphs, a fundamentally different input modality from image-based approaches.}
\label{tab:validation_results}
\end{table}

Detailed performance breakdowns are shown in Fig.~\ref{fig:confusion_matrices}.
Considering the VQA-baseline, Fig.~\ref{fig:confusion_matrices}~(a), the most frequent misclassification occurs between \textit{child’s room} and \textit{bedroom}, as well as between \textit{child’s room} and \textit{studio}. Another notable confusion is between \textit{dressing room} and \textit{bathroom}, where the baseline incorrectly predicts \textit{bathroom} in 21.5\% of \textit{dressing room} cases. Additionally, the model tends to confuse \textit{living room} and \textit{studio}. However, in this case, the baseline is more accurate for \textit{studio} and less accurate for \textit{living room}, which is the opposite pattern of our proposed framework. While the baseline shows slightly less confusion between \textit{bedroom} and \textit{child’s room} than our proposed \methodAcronym~framework, its true positive rate for \textit{child’s room} is considerably lower (52.4\% vs. 65.4\%). Overall, the baseline exhibits lower accuracy for key classes compared to \methodAcronym.

For the \methodAcronym~framework, Fig.~\ref{fig:confusion_matrices}~(b), we highlight the achieved high true positive rates for categories with distinctive objects and spatial arrangements, such as \textit{bathroom}, \textit{bedroom}, \textit{classroom}, \textit{living room}, and \textit{swimming pool}. This suggests that the proposed framework is effective in identifying discriminative triplets that characterize these environments. However, \methodAcronym's most prominent confusion is between \textit{child’s room} and \textit{bedroom}, which is intuitive given their shared core objects (e.g., beds, pillows, windows). A smaller but still noticeable confusion occurs between \textit{living room} and \textit{studio}. Furthermore, we highlight that \methodAcronym~achieved higher accuracy than the baseline in \textit{bedroom}, \textit{child’s room}, \textit{dressing room}, \textit{living room}, and \textit{swimming pool}, showing its advantage in both overall classification and in challenging class pairs. These specific confusion patterns are further explored in the qualitative analysis section.

The performance achieved by our framework in Places8 motivated its application to the sensitive-media scenario, where such discriminative cues could help capture context-specific patterns. To that end, we conducted RCPD experiments using 5-fold cross-validation. Beyond balanced accuracy, we employed recall given its critical importance for sensitive content, as recall directly measures the model's ability to detect CSAI, ensuring such instances are not overlooked.

We began by training the best-performing architecture from Places8 from scratch on binary classification, achieving 72.42\% balanced accuracy and 70.61\% recall. Subsequently, we investigated the potential benefits of transfer learning by fine-tuning the best pre-trained Places8 model. Fine-tuning only the classification head yielded 71.28\% balanced accuracy and 71.44\% recall, while fine-tuning the entire network improved to 73.25\% balanced accuracy and 73.27\% recall. These results suggest a significant domain shift between Places8 indoor scenes and the RCPD context, making naive transfer learning suboptimal. Finally, new hyperparameter optimization with Optuna achieved our best results: \textbf{74.27\%} balanced accuracy and \textbf{76.55\%} recall, underscoring the importance of architectural adaptation for specialized real-world data. Table~\ref{tab:rcpd_results} summarizes these results.\vspace{0.25cm}

\begin{table}[!t]
    \footnotesize
    \centering 
    \begin{tabular}{lcccc}
    \midrule
     & \textbf{Training from scratch} & \textbf{Fine-tuning (head)} & \textbf{Fine-tuning (network)} & \textbf{Optimization}\\
    \midrule
    \textbf{Acc. (\%)} & 72.42 & 71.28 & 73.25 & \textbf{74.27}\\
    \textbf{Recall (\%)} & 70.61 & 71.44 & 73.27 & \textbf{76.55}\\
    \midrule
    \end{tabular}
    \caption{5-fold cross-validation results on the RCPD dataset with our \methodAcronym~framework.}
    \label{tab:rcpd_results}
\end{table}

\begin{table}[t]
\footnotesize
\centering
\setlength{\tabcolsep}{3.25pt}
\begin{tabular}{
>{\raggedright\arraybackslash}p{0.15\linewidth}
>{\raggedright\arraybackslash}p{0.375\linewidth}
>{\raggedright\arraybackslash}p{0.175\linewidth}}
\midrule
\textbf{Class} & \textbf{Top-10 Objects} & \textbf{Top-5 Relations} \\
\midrule
\textit{bathroom} & sink, toilet, handle, door, window, towel, cabinet, tile, shelf, counter & has, on, near, in front of, in \\
\textit{bedroom} & pillow, window, table, chair, bed, door, lamp, room, curtain, [unk\_obj] & has, near, on, in front of, with \\
\textit{child's room} & flower, pillow, [unk\_obj], window, table, chair, shelf, bed, box, bear & has, on, with, near, in front of \\
\textit{classroom} & man, boy, woman, girl, person, table, hair, shirt, chair, [unk\_obj] & on, has, in front of, with, wearing \\
\textit{dressing room} & shelf, door, woman, man, handle, person, window, shirt, bag, [unk\_obj] & on, has, in front of, with, under \\
\textit{living room} & chair, window, table, door, [unk\_obj], room, lamp, shelf, pillow, man & on, has, with, near, in front of \\
\textit{studio} & man, woman, person, shirt, hair, head, window, hand, chair, [unk\_obj] & on, has, in front of, wearing, holding \\
\textit{swimming pool} & window, chair, door, person, table, man, pole, tree, [unk\_obj], building & on, has, near, with, in front of \\
\midrule
\end{tabular}
\caption{Top-10 most influential objects (nodes) and Top-5 most influential relations (edges) per class, ranked by importance. They were aggregated from the GATv2 model's attention weights across all correct predictions in the validation set. The [unk\_obj] token represents objects not recognized in the vocabulary.}
\label{tab:influential_labels}
\end{table}
    
\noindent \textbf{Qualitative Analysis.} Beyond the quantitative metrics, our framework provides inherent explainability through Scene Graphs with Graph Attention Networks. By analyzing the attention coefficients computed by the GATv2 layers, we can move beyond simply assessing accuracy and begin to understand the model's decision-making process. This analysis allows us to pinpoint which objects and relationships contributed most to a given prediction.

First, we performed an aggregated analysis to understand the general patterns learned for each class. By accumulating the attention scores across all correctly classified images, we identified the most important features for each scene category (Table~\ref{tab:influential_labels}). Results confirm the model learns intuitive, human-understandable patterns. For example, \textit{bathroom} is strongly characterized by objects like \texttt{sink}, \texttt{toilet}, and \texttt{towel}. Interestingly, for \textit{classroom}, the model learned that the presence of people (\texttt{man}, \texttt{boy}, and \texttt{woman}) was a more reliable indicator than specific furniture, which can overlap with scenes like \textit{studio} or \textit{living room}.

\begin{figure}[t]
   \centering
    \begin{tabular}{cc}
    \includegraphics[width=0.44\linewidth]{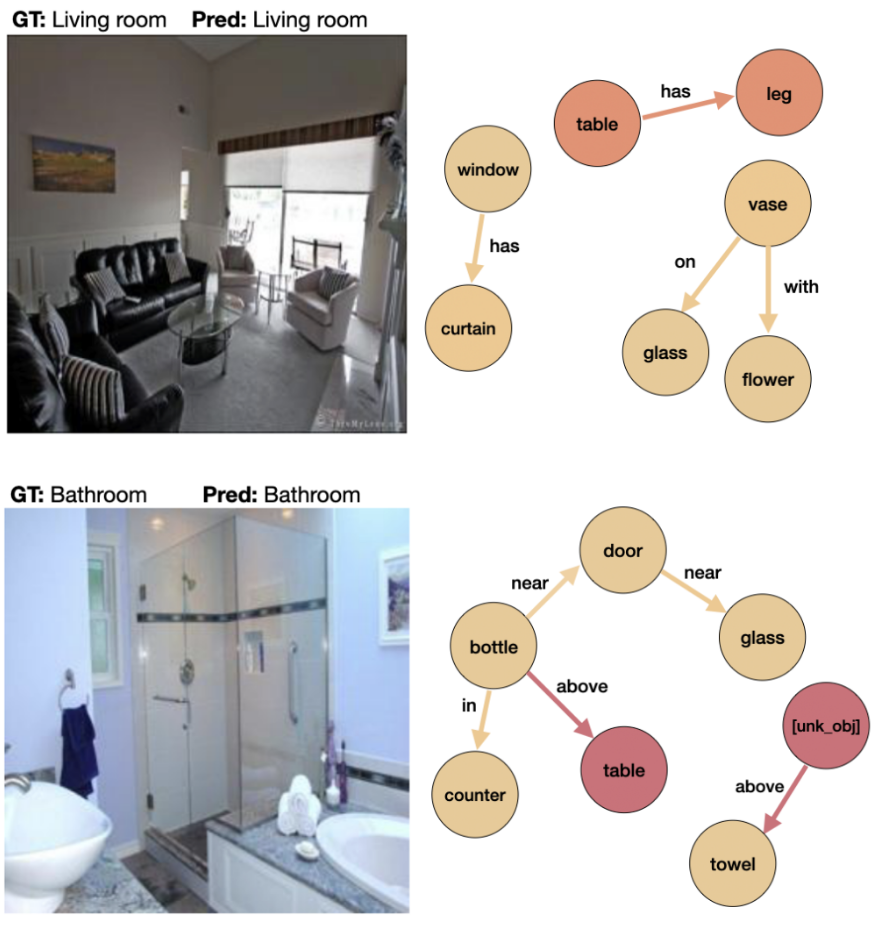} &
    \includegraphics[width=0.53\linewidth]{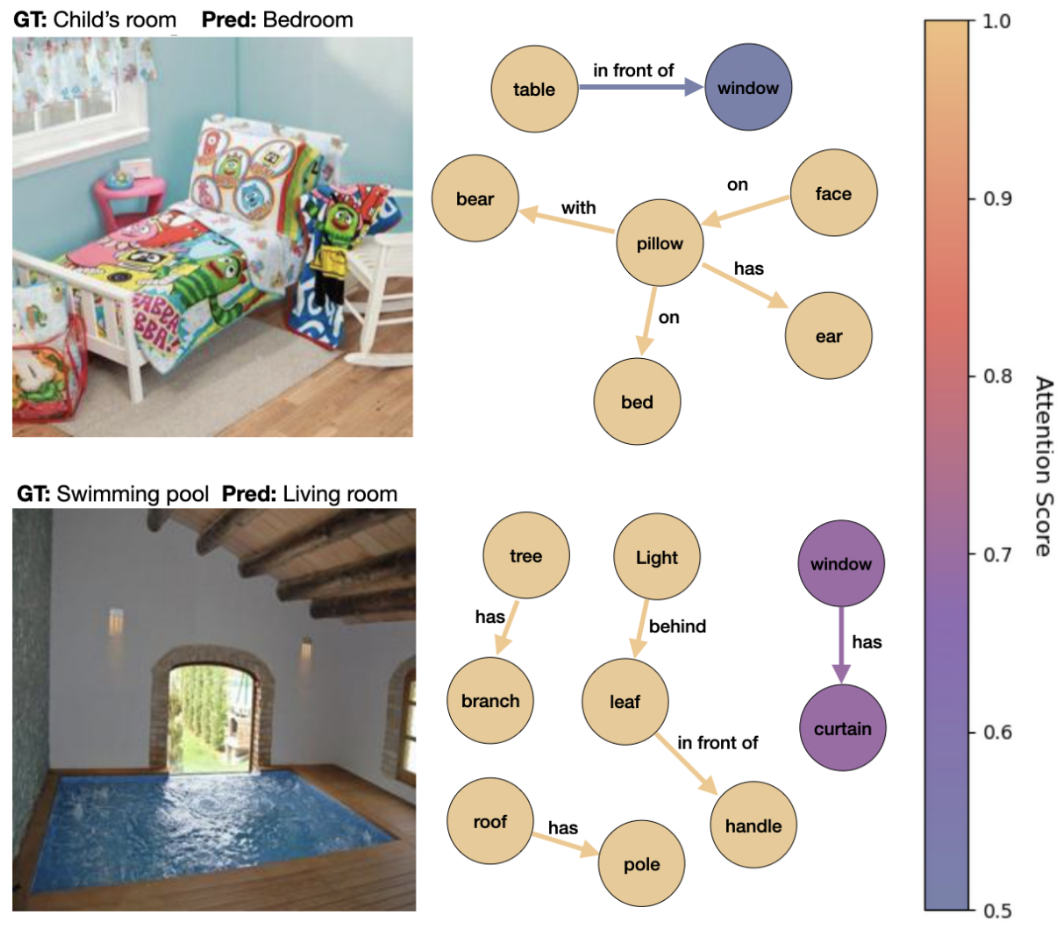} \\
    (a) Correct predictions & (b) Wrong predictions \\\\
    \end{tabular}
   \caption{Qualitative results of \methodAcronym~on  Places8. Column (a) shows correctly classified scenes, while column (b) shows misclassifications.  Each image includes its scene graph with GATv2 attention scores for nodes and edges. These visualizations showcase the model's ability to identify key semantic components for correct predictions and provide transparent analysis of failure cases, including confusion between similar scenes.}
   \label{fig:qualitative_error}
\end{figure}

The attention mechanism effectively diagnoses failure modes. Fig.~\ref{fig:qualitative_error} shows four predictions (two correct, two errors) with SG subsets and attention scores. 
In the correct cases (Fig.~\ref{fig:qualitative_error} (a)), the model assigns higher attention to class-defining objects and their incident relations (lighter color bar tones), e.g., \texttt{window}/\texttt{table} for \emph{living room} and \texttt{sink}/\texttt{towel}/ \texttt{counter} for \emph{bathroom}. Attention also propagates to neighbors of important nodes, reflecting the model's focus on triplets relating to high-attention objects.

In Fig.~\ref{fig:qualitative_error} (b), we observe a typical source of error: upstream SGG hallucinations. The detector incorrectly inserts a \texttt{curtain} next to a \texttt{window}, yielding the high-attention triplet 
$\langle\texttt{window},\texttt{has},\texttt{curtain}\rangle$, which steers the classifier toward \emph{living room} contexts. 
Thresholding SGG triplets by their confidence  did not improve accuracy, as erroneous triplets often receive high scores and survive filtering. This suggests robustness depends more on improving SGG quality and vocabulary than on simple score pruning.

In the first case of Fig.~\ref{fig:qualitative_error}~(b), the model confuses \emph{child’s room} with \emph{bedroom}, a frequent ambiguity in Places8. 
Although the graph includes cues such as \texttt{bear}, \texttt{face}, and \texttt{pillow}, the closed vocabulary cannot express concepts like \texttt{toy}, \texttt{cartoon}, or other child-specific attributes, limiting the semantic signal available to GATv2. Moving to an open-vocabulary SGG or enriching the label space with attributes (e.g., \texttt{toy}, \texttt{crib}, \texttt{cartoon}~\texttt{pattern}) should help disambiguate these classes by capturing the semantics that distinguish a \textit{child’s room} from a  \textit{bedroom}.

In collaboration with law enforcement, we performed the same analysis for CSAI classification. Table~\ref{tab:influential_labels_rcpd} showcases the most important objects and relationships for discerning between categories, enabling nuanced scene understanding even without directly viewing images. For detecting CSAI, \texttt{hand} was the most important object with \texttt{holding} and \texttt{near} as key relationships. This aligns with what expert law enforcement agents may identify~during manual screening, indicating the model's capability to assign attention where most relevant.

\begin{table}[t]
\footnotesize
\centering
\setlength{\tabcolsep}{3.25pt}
\begin{tabular}{
>{\raggedright\arraybackslash}p{0.085\linewidth}
>{\raggedright\arraybackslash}p{0.55\linewidth}
>{\raggedright\arraybackslash}p{0.3\linewidth}}
\midrule
\textbf{Class} & \textbf{Top-10 Objects} & \textbf{Top-5 Relations} \\
\midrule
\textit{CSAI} & hand, woman, head, girl, boy, [unk\_obj], man, person, arm, leg & has, on, near, holding, in front of \\
%\midrule
\textit{Not CSAI} & woman, girl, [unk\_obj], hand, head, hair, man, shirt, boy, leg
& has, on, behind, near, wearing \\
\midrule
\end{tabular}
\caption{Top-10 most influential objects and \text{Top-5} most influential relations per category, ranked by importance, for the RCPD dataset.}
\label{tab:influential_labels_rcpd}
\end{table}

For each evaluated image, a large set of object relations was generated. Although the relation confidence was not the analysis focus, we observed that among the most relevant relations (considering both score and the top-10 most influential classes for  CSAI classification), some were consistent with the actual image content. However, we also identified  implausible or semantically inconsistent relations. This reflects domain shift and the presence of  close-ups and task-specific CSAI content containing visual elements that are presumably absent from the base vocabulary used for object and relation tokenization, highlighting natural limitations in visual element coverage.

\begin{table}[t!]
\footnotesize
\centering
\begin{tabular}{lcl}
\midrule
\textbf{Category} & \textbf{Acc. (\%)} & \textbf{Description} \\
\midrule
CSAI        & 76.97 & Child sexual abuse imagery \\
Not CSAI -- \textit{child}       & 74.29 & Images containing children \\
Not CSAI -- \textit{adult}       & 77.34 & Images containing adults \\
Not CSAI -- \textit{suspicious}  & 72.55 & Children/adolescents in underwear, swimwear, or shirtless \\
Not CSAI -- \textit{pornography}        & 72.41 & Pornographic content \\
Not CSAI -- \textit{normal}      & 84.62 & No nudity \\
CSAI and Not CSAI -- \textit{global}      & 76.69 & Overall dataset \\
\midrule
\end{tabular}
\caption{Accuracy of \methodAcronym \ in CSAI classification on RCPD, reported by image category. While the task remains the same, results are broken down by image content.}
\label{tab:rcpd_quantitative_anal}
\end{table}

\begin{table}[t!]
\footnotesize
    \centering
    \setlength{\tabcolsep}{3.2pt}
    \begin{tabular}{lcccc}
        \midrule
        \textbf{Step} & \textbf{Model} & \textbf{\#Images} & \textbf{Energy (kWH)} & \textbf{CO$_2$-eq (kg)} \\
        \midrule
        SGG & Pix2Grp~\cite{li2024pixels} & 407,640 & 1.0136 & 0.3960\\ 
        %\midrule
        GNN & GATv2~\cite{brody2021attentive} & 364,806 & 1.3520 & 0.1330\\
        \midrule
    \end{tabular}
        \caption{Average energy consumption and equivalent CO$_2$ emissions (CO$_2$-eq) for each step of our method.}
    \label{tab:energy_consumption}
\end{table}

Table~\ref{tab:rcpd_quantitative_anal} shows model accuracy for CSAI classification by image category. Accuracy varies across subsets, with the \textit{normal} category reaching 84.62\%, well above the global average, showing effective classification when no sexual or suggestive content is present. In contrast, performance drops to around 72\% for the more challenging \textit{suspicious} and \textit{pornography} categories, which involve higher visual variability and ambiguous boundaries. For \textit{CSAI}, results are comparable to the \textit{global} accuracy (76.69\%), indicating that despite object-relation vocabulary limitations and the presence of many irrelevant relations, the model successfully exploits the most informative ones to produce adequate final decisions.\vspace{0.25cm}

\noindent \textbf{Environmental Impact Analysis.} 
We also assess the environmental impact of our experiments by quantifying the energy consumption and carbon footprint. To that end, we employed the \textit{CodeCarbon}~tool~\cite{benoit:2024}, which tracks the power usage of our hardware configuration and estimates the carbon emissions generated during the computational processes involved in SGG and GNN training. Table~\ref{tab:energy_consumption} details energy consumption metrics.

\section{Conclusions}

In this work, we introduced \methodAcronym, a novel framework for indoor scene classification that leverages the semantic and the structure of Scene Graphs. Using a Graph Attention Network, our method achieves state-of-the-art 81.27\% balanced accuracy on Places8. In collaboration with law enforcement, we  evaluated \methodAcronym~on real-world CSAI datasets, obtaining 74.27\% balanced accuracy for CSAI classification, showcasing practical utility in digital forensics. The framework's primary strengths are its inherent explainability through attention weight analysis for error diagnosis, and privacy-preserving architecture suitable for sensitive applications like CSAI analysis.

While promising, performance depends on upstream Scene Graph Generation quality and is constrained by closed-set vocabulary. Future work will focus on integrating  advanced, open-vocabulary SGG models and enriching the graph's node and edge features to overcome these limitations.

\section*{Acknowledgments}

This work is partially funded by FAPESP 2023/12086-9, FAEPEX/UNICAMP 2597/23, and the Serrapilheira Institute R-2011-37776. Artur~Barros (2024/09372-2), Carlos~Caetano (2024/01210-3), and Sandra~Avila (2023/12865-8, 2020/09838-0, 2013/08293-7) are also funded by FAPESP. Sandra~Avila is also funded by H.IAAC 01245.003479/2024-10 and CNPq 316489/2023-9.

\bibliography{references}

\begin{thebibliography}{33}
\providecommand{\natexlab}[1]{#1}
\providecommand{\url}[1]{\texttt{#1}}
\expandafter\ifx\csname urlstyle\endcsname\relax
  \providecommand{\doi}[1]{doi: #1}\else
  \providecommand{\doi}{doi: \begingroup \urlstyle{rm}\Url}\fi

\bibitem[Belmecheri et~al.(2025)Belmecheri, Gotlieb, Lazaar, and Spieker]{Belmecheri:2025:Explainable}
Nassim Belmecheri, Arnaud Gotlieb, Nadjib Lazaar, and Helge Spieker.
\newblock Explainable scene understanding with qualitative representations and graph neural networks.
\newblock In \emph{IEEE Intelligent Vehicles Symposium}, 2025.

\bibitem[Brody et~al.(2022)Brody, Alon, and Yahav]{brody2021attentive}
Shaked Brody, Uri Alon, and Eran Yahav.
\newblock How attentive are graph attention networks?
\newblock In \emph{International Conference on Learning Representations (ICLR)}, 2022.

\bibitem[Carlini et~al.(2021)Carlini, Tram{\`e}r, Wallace, Jagielski, {Herbert-Voss}, Lee, Roberts, Brown, Song, Erlingsson, Oprea, and Raffel]{Carlini2021ExtractingTrainingData}
Nicholas Carlini, Florian Tram{\`e}r, Eric Wallace, Matthew Jagielski, Ariel {Herbert-Voss}, Katherine Lee, Adam Roberts, Tom Brown, Dawn Song, {\'U}lfar Erlingsson, Alina Oprea, and Colin Raffel.
\newblock Extracting {{Training Data}} from {{Large Language Models}}.
\newblock In \emph{30th USENIX security symposium (USENIX Security 21)}, 2021.

\bibitem[Chen et~al.(2020)Chen, Song, Zeng, and Jiang]{chen2020scene}
Gongwei Chen, Xinhang Song, Haitao Zeng, and Shuqiang Jiang.
\newblock Scene recognition with prototype-agnostic scene layout.
\newblock \emph{IEEE Transactions on Image Processing (TIP)}, 29:\penalty0 5877--5888, 2020.

\bibitem[Chiang et~al.(2023)Chiang, Li, Lin, Sheng, Wu, Zhang, Zheng, Zhuang, Zhuang, Gonzalez, Stoica, and Xing]{Chiang:2023:Vicuna}
Wei-Lin Chiang, Zhuohan Li, Zi~Lin, Ying Sheng, Zhanghao Wu, Hao Zhang, Lianmin Zheng, Siyuan Zhuang, Yonghao Zhuang, Joseph~E. Gonzalez, Ion Stoica, and Eric~P. Xing.
\newblock Vicuna: An open-source chatbot impressing gpt-4 with 90\%* chatgpt quality, March 2023.
\newblock URL \url{https://lmsys.org/blog/2023-03-30-vicuna/}.

\bibitem[Coelho et~al.(2024)Coelho, Ribeiro, Macedo, dos Santos, and Avila]{coelho2024transformers}
Thamiris Coelho, Leo S.~F. Ribeiro, Jo{\~a}o Macedo, Jefersson~A dos Santos, and Sandra Avila.
\newblock Transformers-based few-shot learning for scene classification in child sexual abuse imagery.
\newblock In \emph{IEEE Conference on Graphics, Patterns and Images (SIBGRAPI)}, pages 8--14, 2024.

\bibitem[Coelho et~al.(2025)Coelho, Ribeiro, Macedo, dos Santos, and Avila]{Coelho:2025}
Thamiris Coelho, Leo S.~F. Ribeiro, Jo\~{a}o Macedo, Jefersson~A. dos Santos, and Sandra Avila.
\newblock Minimizing risk through minimizing model-data interaction: A protocol for relying on proxy tasks when designing child sexual abuse imagery detection models.
\newblock In \emph{ACM Conference on Fairness, Accountability, and Transparency (FAccT)}, pages 1543--1553, 2025.

\bibitem[Courty et~al.(2024)Courty, Schmidt, Luccioni, Goyal-Kamal, MarionCoutarel, Feld, Lecourt, LiamConnell, Saboni, et~al.]{benoit:2024}
Benoit Courty, Victor Schmidt, Sasha Luccioni, Goyal-Kamal, MarionCoutarel, Boris Feld, Jérémy Lecourt, LiamConnell, Amine Saboni, et~al.
\newblock mlco2/codecarbon: v2.4.1, 2024.
\newblock URL \url{https://doi.org/10.5281/zenodo.11171501}.

\bibitem[Fan et~al.(2022)Fan, Liu, Chen, Ramesh, and Xiang]{Fan:2022:SRRN}
Kanglong Fan, Wei Liu, Xiaowen Chen, Bharath Ramesh, and Cheng Xiang.
\newblock An interpretable scene understanding framework via graph learning.
\newblock \emph{SSRN 4238333}, 2022.

\bibitem[Jung et~al.(2023)Jung, Kim, Kim, and Cho]{Jung:CVPR:2023}
Deunsol Jung, Sanghyun Kim, Won~Hwa Kim, and Minsu Cho.
\newblock Devil's on the edges: Selective quad attention for scene graph generation.
\newblock In \emph{IEEE/CVF Conference on Computer Vision and Pattern Recognition (CVPR)}, pages 18664--18674, 2023.

\bibitem[Kingma and Ba(2014)]{Kingma2015ADAM}
Diederik Kingma and Jimmy Ba.
\newblock Adam: {{A}} method for stochastic optimization.
\newblock In \emph{International Conference on Learning Representations (ICLR)}, 2014.

\bibitem[Kloess et~al.(2021)Kloess, Woodhams, and Hamilton-Giachritsis]{Kloess:2021}
Juliane~A. Kloess, Jessica Woodhams, and Catherine~E. Hamilton-Giachritsis.
\newblock The challenges of identifying and classifying child sexual exploitation material: Moving towards a more ecologically valid pilot study with digital forensics analysts.
\newblock \emph{Child Abuse \& Neglect}, 118:\penalty0 105166, 2021.

\bibitem[Krishna et~al.(2017)Krishna, Zhu, Groth, Johnson, Hata, Kravitz, Chen, Kalantidis, Li, Shamma, et~al.]{Krishna:IJCV:2017}
Ranjay Krishna, Yuke Zhu, Oliver Groth, Justin Johnson, Kenji Hata, Joshua Kravitz, Stephanie Chen, Yannis Kalantidis, Li-Jia Li, David~A Shamma, et~al.
\newblock Visual genome: Connecting language and vision using crowdsourced dense image annotations.
\newblock \emph{International Journal of Computer Vision (IJCV)}, 123\penalty0 (1):\penalty0 32--73, 2017.

\bibitem[Li et~al.(2024)Li, Zhang, Lin, Chen, and He]{li2024pixels}
Rongjie Li, Songyang Zhang, Dahua Lin, Kai Chen, and Xuming He.
\newblock From pixels to graphs: Open-vocabulary scene graph generation with vision-language models.
\newblock In \emph{IEEE/CVF Conference on Computer Vision and Pattern Recognition (CVPR)}, pages 28076--28086, 2024.

\bibitem[Liu et~al.(2023)Liu, Li, Wu, and Lee]{Liu:2023:LLaVA}
Haotian Liu, Chunyuan Li, Qingyang Wu, and Yong~Jae Lee.
\newblock Visual instruction tuning.
\newblock In \emph{Advances in Neural Information Processing Systems (NeurIPS)}, pages 34892--34916, 2023.

\bibitem[Liu et~al.(2024)Liu, Li, Li, Li, Zhang, Shen, and Lee]{Liu:2024:LLaVA1.6-NeXT}
Haotian Liu, Chunyuan Li, Yuheng Li, Bo~Li, Yuanhan Zhang, Sheng Shen, and Yong~Jae Lee.
\newblock Llava-next: Improved reasoning, ocr, and world knowledge, January 2024.
\newblock URL \url{https://llava-vl.github.io/blog/2024-01-30-llava-next/}.

\bibitem[Macedo et~al.(2018)Macedo, Costa, and {A. dos Santos}]{Macedo2018BenchmarkMethodologyChild}
Joao Macedo, Filipe Costa, and Jefersson {A. dos Santos}.
\newblock A benchmark methodology for child pornography detection.
\newblock In \emph{{{Conference}} on {{Graphics}}, {{Patterns}} and {{Images}} ({{SIBGRAPI}})}, pages 455--462, 2018.

\bibitem[Patel et~al.(2020)Patel, Dabhi, and Prajapati]{patel2020survey}
Tanvi~A Patel, Vipul~K Dabhi, and Harshadkumar~B Prajapati.
\newblock Survey on scene classification techniques.
\newblock In \emph{IEEE International Conference on Advanced Computing and Communication Systems (ICACCS)}, pages 452--458, 2020.

\bibitem[Qiu et~al.(2021)Qiu, Yang, Wang, and Tao]{Qiu2021SceneEssence}
Jiayan Qiu, Yiding Yang, Xinchao Wang, and Dacheng Tao.
\newblock Scene {{Essence}}.
\newblock In \emph{IEEE/CVF Conference on Computer Vision and Pattern Recognition (CVPR)}, pages 8318--8329, 2021.

\bibitem[Quattoni and Torralba(2009)]{Quattoni2009MITIndoorSR}
Ariadna Quattoni and Antonio Torralba.
\newblock Recognizing indoor scenes.
\newblock In \emph{IEEE/CVF Conference on Computer Vision and Pattern Recognition (CVPR)}, pages 413--420, 2009.

\bibitem[Radford et~al.(2021)Radford, Kim, Hallacy, Ramesh, Goh, Agarwal, Sastry, Askell, Mishkin, Clark, et~al.]{Radford:2021:CLIP}
Alec Radford, Jong~Wook Kim, Chris Hallacy, Aditya Ramesh, Gabriel Goh, Sandhini Agarwal, Girish Sastry, Amanda Askell, Pamela Mishkin, Jack Clark, et~al.
\newblock Learning transferable visual models from natural language supervision.
\newblock In \emph{International Conference on Machine Learning (ICML)}, pages 8748--8763, 2021.

\bibitem[Sajjadi et~al.(2023)Sajjadi, Mahendran, Kipf, Pot, Duckworth, Lu\v{c}i\'c, and Greff]{Sajjadi:2023:CVPR}
Mehdi S.~M. Sajjadi, Aravindh Mahendran, Thomas Kipf, Etienne Pot, Daniel Duckworth, Mario Lu\v{c}i\'c, and Klaus Greff.
\newblock {RUST}: Latent neural scene representations from unposed imagery.
\newblock In \emph{IEEE/CVF Conference on Computer Vision and Pattern Recognition (CVPR)}, pages 17297--17306, 2023.

\bibitem[Song and Ma(2023)]{Song:2023:IJCNN}
Chuanxin Song and Xin Ma.
\newblock Srrm: Semantic region relation model for indoor scene recognition.
\newblock In \emph{International Joint Conference on Neural Networks (IJCNN)}, 2023.

\bibitem[Tang et~al.(2019)Tang, Zhang, Wu, Luo, and Liu]{Tang_2019_CVPR}
Kaihua Tang, Hanwang Zhang, Baoyuan Wu, Wenhan Luo, and Wei Liu.
\newblock Learning to compose dynamic tree structures for visual contexts.
\newblock In \emph{IEEE/CVF Conference on Computer Vision and Pattern Recognition (CVPR)}, pages 9002--9011, 2019.

\bibitem[Tong et~al.(2024)Tong, Brown, Wu, Woo, Middepogu, Akula, Yang, Yang, Iyer, Pan, Wang, Fergus, LeCun, and Xie]{Tong:2024:NeurIPS}
Shengbang Tong, Ellis Brown, Penghao Wu, Sanghyun Woo, Manoj Middepogu, Sai~Charitha Akula, Jihan Yang, Shusheng Yang, Adithya Iyer, Xichen Pan, Austin Wang, Rob Fergus, Yann LeCun, and Saining Xie.
\newblock Cambrian-1: a fully open, vision-centric exploration of multimodal llms.
\newblock In \emph{Advances on Neural Information Processing Systems (NeurIPS)}, 2024.

\bibitem[Valois et~al.(2025)Valois, Macedo, Ribeiro, dos Santos, and Avila]{valois2025leveraging}
Pedro H.~V. Valois, Jo{\~a}o Macedo, Leo S.~F. Ribeiro, Jefersson~A dos Santos, and Sandra Avila.
\newblock Leveraging self-supervised learning for scene classification in child sexual abuse imagery.
\newblock \emph{Forensic Science International: Digital Investigation}, 53:\penalty0 301918, 2025.

\bibitem[Velickovic et~al.(2018)Velickovic, Cucurull, Casanova, Romero, Li{\`o}, and Bengio]{Velickovic2018GGraphAttentionNetworks}
Petar Velickovic, Guillem Cucurull, Arantxa Casanova, Adriana Romero, Pietro Li{\`o}, and Yoshua Bengio.
\newblock Graph attention networks.
\newblock In \emph{International Conference on Learning Representations (ICLR)}, 2018.

\bibitem[Wang et~al.(2023)Wang, Dai, Chen, Huang, Li, Zhu, Hu, Lu, Lu, Li, et~al.]{wang2023internimage}
Wenhai Wang, Jifeng Dai, Zhe Chen, Zhenhang Huang, Zhiqi Li, Xizhou Zhu, Xiaowei Hu, Tong Lu, Lewei Lu, Hongsheng Li, et~al.
\newblock Internimage: Exploring large-scale vision foundation models with deformable convolutions.
\newblock In \emph{IEEE/CVF Conference on Computer Vision and Pattern Recognition (CVPR)}, pages 14408--14419, 2023.

\bibitem[Xu et~al.(2017)Xu, Zhu, Choy, and Fei-Fei]{Xu:CVPR:2017}
Danfei Xu, Yuke Zhu, Christopher~B. Choy, and Li~Fei-Fei.
\newblock Scene graph generation by iterative message passing.
\newblock In \emph{IEEE/CVF Conference on Computer Vision and Pattern Recognition (CVPR)}, 2017.

\bibitem[Yue et~al.(2024)Yue, Ni, Zhang, Zheng, Liu, Zhang, Stevens, Jiang, Ren, Sun, Wei, Yu, Yuan, Sun, Yin, Zheng, Yang, Liu, Huang, Sun, Su, and Chen]{Yue:2023:CVPR}
Xiang Yue, Yuansheng Ni, Kai Zhang, Tianyu Zheng, Ruoqi Liu, Ge~Zhang, Samuel Stevens, Dongfu Jiang, Weiming Ren, Yuxuan Sun, Cong Wei, Botao Yu, Ruibin Yuan, Renliang Sun, Ming Yin, Boyuan Zheng, Zhenzhu Yang, Yibo Liu, Wenhao Huang, Huan Sun, Yu~Su, and Wenhu Chen.
\newblock {MMMU}: A massive multi-discipline multimodal understanding and reasoning benchmark for expert agi.
\newblock In \emph{IEEE/CVF Conference on Computer Vision and Pattern Recognition (CVPR)}, 2024.

\bibitem[Yue et~al.(2025)Yue, Zheng, Ni, Wang, Zhang, Tong, Sun, Yu, Zhang, Sun, Su, Chen, and Neubig]{Yue:2025:ACL}
Xiang Yue, Tianyu Zheng, Yuansheng Ni, Yubo Wang, Kai Zhang, Shengbang Tong, Yuxuan Sun, Botao Yu, Ge~Zhang, Huan Sun, Yu~Su, Wenhu Chen, and Graham Neubig.
\newblock {MMMU-Pro}: A more robust multi-discipline multimodal understanding benchmark.
\newblock In \emph{Annual Meeting of the Association for Computational Linguistics (ACL)}, pages 15134--15186, 2025.

\bibitem[Zhou et~al.(2014)Zhou, Lapedriza, Xiao, Torralba, and Oliva]{zhou2014}
Bolei Zhou, Agata Lapedriza, Jianxiong Xiao, Antonio Torralba, and Aude Oliva.
\newblock Learning deep features for scene recognition using places database.
\newblock In \emph{Advances in Neural Information Processing Systems (NeurIPS)}, pages 487--495, 2014.

\bibitem[Zhou et~al.(2017)Zhou, Lapedriza, Khosla, Oliva, and Torralba]{Zhou2017PlacesSR}
Bolei Zhou, Agata Lapedriza, Aditya Khosla, Aude Oliva, and Antonio Torralba.
\newblock Places: {{A}} 10 million image database for scene recognition.
\newblock \emph{IEEE Transactions on Pattern Analysis and Machine Intelligence (TPAMI)}, pages 1452--1464, 2017.

\end{thebibliography}
\end{document}